\documentclass[11pt]{article}
\usepackage{graphicx} 
\usepackage{amsmath}
\DeclareMathOperator*{\argmax}{argmax}
\usepackage{amssymb}
\usepackage{color,soul}
\usepackage{appendix}
\usepackage{float}
\usepackage{cancel}
\usepackage{hyperref}
\usepackage{multirow}
\usepackage[dvipsnames]{xcolor}
\usepackage[most]{tcolorbox}
\usepackage[a4paper]{geometry}
\usepackage{setspace}
\newtcolorbox[auto counter]{assumptionbox}[2][]{%
    colback=blue!5!white,
    colframe=blue!75!black,
    colbacktitle=purple!20!blue!75,
    coltitle=white,
    title=Box~\thetcbcounter\ - #2,
    fonttitle=\bfseries,
    enhanced,
    frame hidden,
    drop fuzzy shadow,
    #1
}

\title{Bio-Inspired Artificial Neural Networks based on Predictive Coding}
\author{
    Davide Casnici,
    Charlotte Frenkel*,
    Justin Dauwels*\\
    \small{*Equal contributions}
}
\date{}

\begin{document}

\maketitle

\let\thefootnote\relax\footnotetext{%
\noindent
This work has been submitted to the IEEE for possible publication.\\
Copyright may be transferred without notice, after which this version may no longer be accessible.\\
Davide Casnici, Charlotte Frenkel and Justin Dauwels are with the Microelectronics Department (EEMCS Faculty), Delft University of Technology, 2628 CD Delft, Netherlands (e-mail: d.casnici@tudelft.nl).
}

Backpropagation (BP) of errors is currently the backbone training algorithm for artificial neural networks (ANNs). It works by updating the network weights through gradient descent to minimize the value of a \textit{loss function}, which represents the mismatch between the network's prediction and the desired output. BP relies on the chain rule from calculus to propagate the loss gradient backward through the network's hierarchy, allowing each weight to be efficiently and precisely updated based on its contribution to the output error.
However, this process constrains the weight updates at every layer to rely on a \textit{global} error signal generated at the extreme of the hierarchy. 
In contrast, the Hebbian model of synaptic plasticity in the brain states that weight updates should be \textit{local}, determined only by the activity of pre- and post-synaptic neurons. According to Hebb's model, it is therefore unlikely that biological brains directly implement BP. Recently, an alternative algorithm for training ANNs called Predictive Coding (PC) is gaining interest, appearing as a more biologically plausible alternative that updates the network weights using only local information. 
Originating from P. Elias’s 1950s work on signal compression \cite{elias1955predictiveI}, PC was later proposed in neuroscience as a model of the visual cortex by Rao and Ballard \cite{rao1999predictive}. Successively, K. Friston formalized it under the free energy principle (FEP) \cite{friston2005theory, friston2009predictive}, grounding PC within the frameworks of Bayesian inference and dynamical systems.  PC weight updates rely only on pre- and post-synaptic information, eliminating BP's dependence on a global error signal. Moreover, it theoretically provides features beyond those of standard BP, such as the ability to automatically scale each gradient based on the associated uncertainty.

This Lecture Notes column offers a novel, tutorial-style introduction to PC, focusing on its formulation, derivation, and connections to well-established optimization and signal processing algorithms such as BP and the Kalman Filter (KF).  It aims to provide accessible support to the existing literature, guiding readers from the mathematical foundations underlying PC to its practical implementation, including computational examples in Python using the PyTorch framework.

\section*{Relevance}
While BP has enabled significant advances in training large-scale models, it lacks biological plausibility. Weight updates in early layers of the network rely on error signals generated far away in the hierarchy, violating locality principle central to Hebbian plasticity. PC offers a biologically plausible alternative, enabling parameter updates using only locally available information. Notably, recent studies have shown that, under certain conditions, PC can approximate or even exactly match BP gradients \cite{whittington2017approximation, millidge2022predictive, song2020can}. Furthermore, PC has been framed as Bayesian filtering and shown to approximate the closed-form Kalman Filter (KF) solution when recurrent connections are included \cite{millidge2021neural, millidge2024predictive}. These properties make PC a promising algorithm for next-generation biologically plausible ANNs, linking biological plausibility to already well-established artificial intelligence (AI) and signal processing algorithms.

In this Lecture Notes column, we will guide step-by-step readers through the mathematical foundations of PC, providing clear and intuitive explanations of all required tools and key concepts. We will also briefly cover its relations with BP and KF algorithms, and conclude by discussing some trade-offs with a computational example.

\section*{Prerequisites}
The content of this Lecture Notes column is meant to be as self-contained as possible; however, readers must be familiar with calculus, linear algebra, probability, and Bayesian statistics. A basic knowledge of variational inference and KF is helpful but not essential.

\section*{Problem statement and solution}

\subsection*{Problem statement}
A key question in the fields of neuroscience and AI is how the brain processes and makes sense of sensory signals, giving rise to perceptions. When sensing some stimuli, the brain has to infer and interpret the physical setting in the real world that has produced them, resulting in what we experience as perception. This task is particularly complex when considering that synaptic changes in the brain rely exclusively on local plasticity. Therefore, how could this perception problem be solved by relying only on local learning rules, and what kind of connections might the resulting mathematical model share with AI and system identification algorithms?

\subsection*{Predictive Coding approach}
Helmholtz's principle states that when there is a significant deviation from randomness, structure becomes apparent to us, and perception is thus an unconscious inference process about statistical regularities of sensory stimuli. PC suggests that the brain represents these statistical regularities through neural activity, aiming to predict and infer the causes of unforeseen stimuli. The brain is therefore assumed to employ a generative model of the world, and to infer the causes of sensory stimuli by inverting the generative process \cite{friston2009predictive}. Here, ``inversion" refers to the process of inferring causes from stimuli rather than generating stimuli from causes. These causes are encoded as neural activity at varying abstraction levels, such as edges or more complex shapes. For example, while reading this manuscript, the brain infers the most likely physical state of the world that causes the light detected by the eyes (the black shapes of letters on a white background). Its predictions are compared with the actual sensory input, and any mismatch generates error feedback that suppresses inaccurate beliefs. This results in a final neural activity pattern that we experience as the visual perception of this manuscript. Therefore, PC suggests that the visual cortex actively predicts sensory inputs rather than passively processing them. Indeed, PC networks are characterized by top-down prediction signals that predict sensory inputs and bottom-up error signals that transmit the residuals of these predictions, conveying only the unpredicted information \cite{rao1999predictive}.
Having summarized the core idea behind PC, we now introduce the key mathematical concepts essential for the later sections.

\subsection*{Information Measures}
In this section, we will cover the basic mathematical concepts and tools from information theory (IT) that will be the bedrock for later derivations and understanding of PC. 

\begin{figure}[htbp]
    \centering
    \includegraphics[width=0.99\textwidth]{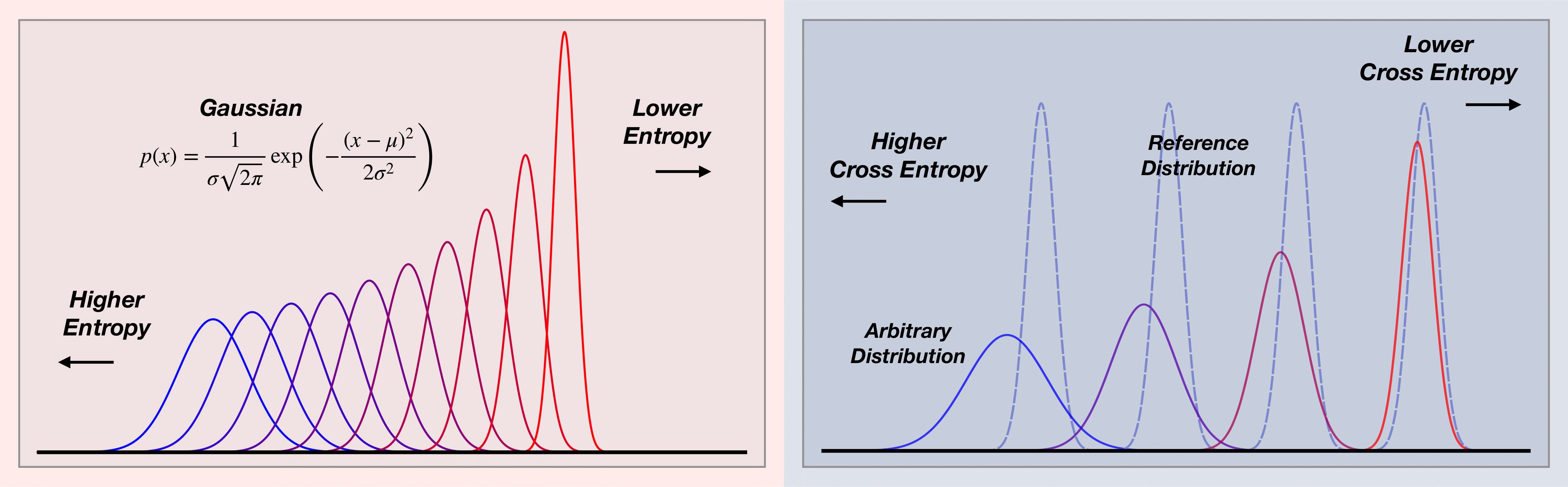}
    \caption{\doublespacing The left pane illustrates how the shape of a Gaussian distribution varies with its entropy, providing an intuitive understanding of this information-theoretic concept. The right pane compares four coloured Gaussian distributions to a dashed-line reference; the closer their means and variances match those of the reference distribution, the lower the cross-entropy, and vice versa.}
    \label{fig:entropy}
\end{figure}

Given a random variable $X$ defined over a discrete support $\mathcal{X} = \{x_1, \dots, x_n\}$, we may be interested in quantifying the surprisal, or information content, $I(x)$ of its realizations. According to Shannon's definition \cite{marsh2013introduction}, the information content of a realization $x \in \mathcal{X}$ of a random variable $X$ with probability mass function $p(x)$, where $X \sim p(x)$ and $p: \mathcal{X} \to [0,1]$, is given by
\begin{equation}
I(x) \triangleq -\log_{2}{p(x)},
\label{eq:1}
\end{equation}
where $p(x)$ denotes the probability of $X$ taking the specific value $x$. While it is usually denoted as $P(X=x)$, for the sake of brevity we will use the simplified notation throughout this manuscript. The $I(\cdot)$ operator allows us to quantify the expected information content of a random variable, referred to as Shannon's \textit{entropy}, defined as

\begin{equation}
H(X) \triangleq E[I(x)] = -\sum_{x \in \mathcal{X}} p\left(x\right)\log_{2}p\left(x\right).
\label{eq:2}
\end{equation}
Its unit depends on the logarithm's base; with base 2, it is measured in bits. Using the natural logarithm, as we will later, quantifies information in nats, where \( 1\,\text{nat} = \log_2 e\,\text{bits} \). The entropy of a random variable can be interpreted as a measure of the uncertainty of its outcomes, or equivalently, as the minimum average number of bits needed to encode a sample from the random variable's distribution losslessly. As shown in the left pane of Figure \ref{fig:entropy}, the more peaked a distribution is around some values, the lower its entropy, and vice versa, since most likely outcomes can be encoded with fewer bits than unlikely ones. For a continuous random variable distributed according to a probability density function \( X \sim f(x), \, x \in \mathcal{X} \subseteq \mathbb{R} \), the summation in \eqref{eq:1} is replaced by an integral and the probability mass function is replaced by the probability density function; thus, the entropy is defined as

\begin{equation}
H(X) \triangleq E[I(X)] = - \int_{\mathcal{X}} f\left(x\right) \log_{2} f\left(x\right) \, dx.
\label{eq:3}
\end{equation}
It is important to highlight that in the continuous case, Shannon's entropy is no longer rigorously defined, and it loses its original interpretation as an information measure. In the discrete case, it is a non-negative quantity representing the expected information content of a random variable. However, in the continuous case, entropy can result in negative values, and this interpretation no longer holds \cite{marsh2013introduction}. Through the manuscript, the entropy of a random variable $X \sim p(x)$ will be denoted as $H(p)$.

Another quantity of interest is the \textit{cross entropy}. It measures the average number of bits needed to encode the outcomes of a random variable distributed according to a reference probability distribution $p$, under another arbitrary probability distribution $q$. Intuitively, it has high values when events that are likely according to the true distribution $p$ are assigned a low probability under the distribution $q$, as shown by the right pane of Figure \ref{fig:entropy}. That is, the cross entropy between two probability distributions $p$ and $q$ defined over the same discrete support $\mathcal{X} = \{x_1, \dots, x_n\}$ is defined as

\begin{equation}
H(p, q) \triangleq -E_p\left[\log_{2} q(x)\right] = - \sum_{x \in \mathcal{X}} p\left(x\right)\log_{2}q\left(x\right).
\label{eq:4}
\end{equation}
The cross entropy can also be expressed in terms of the reference probability distribution's entropy added to the \textit{Kullback–Leibler divergence} ($D_{KL}$) between the reference and arbitrary probability distributions: 

\begin{equation}
H(p, q) = H(p) + D_{\text{KL}}(p \| q).
\label{eq:5}
\end{equation}
The $D_{KL}$ is a measure of dissimilarity between an arbitrary probability distribution and a reference distribution, defined in the discrete case as

\begin{equation}
   D_{\text{KL}}(p\| q) \triangleq \sum_{x \in \mathcal{X}} p(x) \log_2 \left( \frac{p(x)}{q(x)}\right).
   \label{eq:6}
\end{equation}
Both cross entropy and $D_{KL}$ are asymmetric measures, that is $D_{KL}(q||p) \neq D_{KL}(p||q)$ and $H(p,q) \neq H(q,p)$. For continuous random variables, the summation is replaced by an integral in both definitions. From the right pane in Figure \ref{fig:entropy}, we can observe that as the reference distribution and the arbitrary distribution become more similar, their cross entropy decreases, and vice versa.

Now that we have introduced the essential tools from IT, we will cover how they connect to the variational inference (VI) framework, the backbone of PC.

\subsection*{Variational Inference}
In this section, we will cover VI, a technique that allows us to work with complex or intractable probability distributions by approximating them with arbitrary distributions that we can work with. 

\begin{figure}[H]
    \centering
    \includegraphics[width=0.99\textwidth]{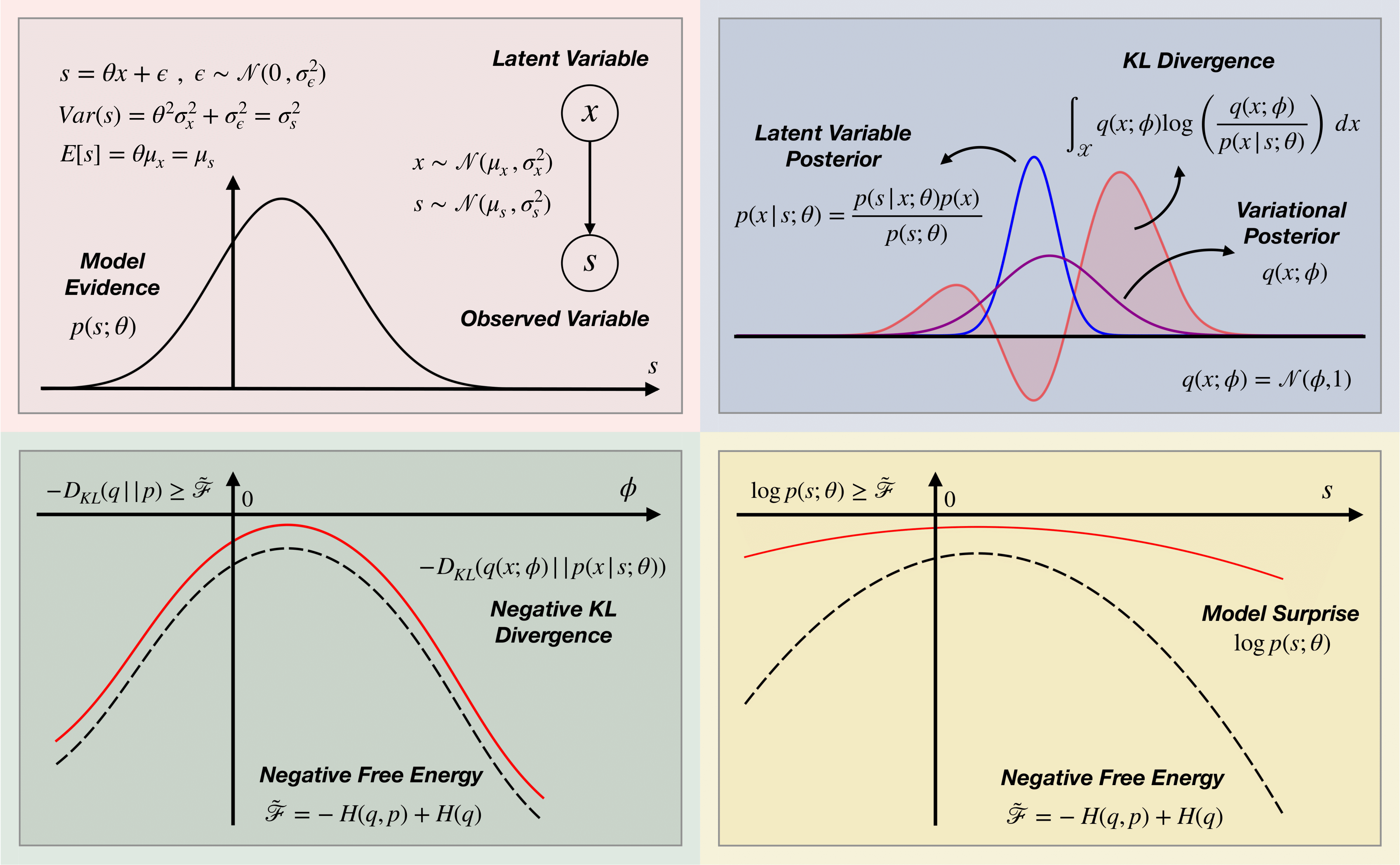}
    \caption{
    \doublespacing The top left pane illustrates the distribution of a random variable conditioned on a latent one, both following Gaussian distributions. The top right illustrates in red the area representing $D_{KL}$ between the real posterior (given observation $s$) and a Gaussian variational posterior. The bottom left compares $D_{KL}$ and the NFE across variational parameters, showing NFE as a lower bound. The bottom right displays model surprise (log-evidence) versus the NFE, showing how the latter acts as a lower bound also for the log-evidence.\color{black}
    }
    \label{fig:dkl}
\end{figure}

When working with Bayesian inference, it is common to encounter integrals that are too complex to be solved both numerically and analytically. For instance, let us suppose we are given an observation $s \in \mathcal{S} \subseteq \mathbb{R}$ randomly generated by a statistical model $p(s; \theta)$, parametrized by some parameter $\theta \in \Theta \subseteq \mathbb{R}$. Assuming this generative model has some latent variable $x \in \mathcal{X} \subseteq \mathbb{R}$ that influences our observation $s$, with this relationship governed by the model parameter, we can represent the joint distribution as $p(s, x; \theta)$. In this case, we might be interested in inferring the most likely value of the latent variable $x$ given the observed data $s$ and the model parameter $\theta$. Using Bayes' theorem, we can compute the posterior distribution of the latent variable given the observed data as

\begin{equation}
   p(x | s; \theta) = \frac{p(s | x; \theta) p(x)}{p(s; \theta)},
   \label{eq:7}
\end{equation}
where \( p(s; \theta) \), known as the model evidence or marginal likelihood, is the normalizing constant obtained by marginalizing over all possible values of \( x \). However, when dealing with high-dimensional multivariate distributions with a large number of latent variables involved, computing the model evidence often becomes intractable. If we assume a latent variable $x$ to be a high-dimensional array of parameters, $x \in \mathcal{X}^{m} \subseteq \mathbb{R}^{m}$, \eqref{eq:7} is written as

\begin{equation}
   p(x | s; \theta) = \frac{p(s | x; \theta) p(x)}{\int_{\mathcal{X}} \dots \int_{\mathcal{X}} p(s | x; \theta) p(x)\,dx_1 \dots dx_m\,},
   \label{eq:8}
\end{equation}
where computing the integral in the denominator may require an intractable amount of computation, making it unfeasible to evaluate the posterior distribution $p(x | s; \theta)$. The core idea of VI is to approximate the intractable posterior distribution with an arbitrary tractable distribution. The latter is referred to as the \textit{variational posterior}, denoted by $q$. To allow for some degrees of flexibility, it is usually parametrized by some parameters $\phi \in \Phi \subseteq \mathbb{R}^{m}$, that can be learnt. To quantify the difference between the posterior and variational distributions $p$ and $q$, we use the previously defined $D_{\mathrm{KL}}$: 

\begin{equation}
   D_{\text{KL}}(q\|p) \triangleq \int_{\mathcal{X}} q(x;\phi) \ln \left( \frac{q(x;\phi)}{p(x | s; \theta)}\right)\, dx\,,
   \label{eq:9}
\end{equation}
where we avoid expanding the integral as in \eqref{eq:8} for brevity. The top-right pane of Figure~\ref{fig:dkl} visually illustrates \eqref{eq:9}, with the red area representing the integrand of the $D_{\mathrm{KL}}$. To minimize the difference between the two distributions, we can modify the parameters of our variational posterior according to the $D_{KL}$ gradient, but this would still require computing the intractable posterior. To overcome this, we can substitute \eqref{eq:7} into \eqref{eq:9} obtaining:

\begin{equation}
   D_{\text{KL}}(q\|p) = \int_{\mathcal{X}} q(x;\phi) \ln \left( \frac{q(x;\phi)}{p(s | x; \theta) p(x)}p(s; \theta)\right)\, dx\,.
   \label{eq:10}
\end{equation}
Continuing by applying the logarithms sum rule, we can develop \eqref{eq:10} as follows to obtain a more familiar equivalent form:
\begin{equation}
\begin{aligned}
   D_{\text{KL}}(q\|p) &= \int_{\mathcal{X}} q(x;\phi) \ln \left( \frac{q(x;\phi)}{p(s,x;\theta)}\right) \, dx + \ln p(s; \theta) \int_{\mathcal{X}} q(x;\phi)  \, dx \\
   &= \int_{\mathcal{X}} q(x;\phi) \ln q(x;\phi) \, dx - \int_{\mathcal{X}} q(x;\phi) \ln p(s, x;\theta) \, dx + \ln p(s; \theta).
\end{aligned}
\label{eq:11}
\end{equation}
This decomposition of the $D_{KL}$ is sometimes referred to as the \textit{energy-entropy decomposition}, due to its similarity with Helmholtz's free energy in statistical physics. By looking closer at \eqref{eq:11}, we can easily recognize two familiar terms: the cross entropy and the entropy. Readers may find the term cross entropy referred to as the `energy' in the FEP or PC literature. Proceeding by applying definitions in \eqref{eq:3} and \eqref{eq:4}, we can write \eqref{eq:11} as

\begin{equation}
   D_{\text{KL}}(q\|p) = -H(q) + H(q, p) + \ln p(s; \theta).
\label{eq:12}
\end{equation}
Since we wish to minimize the divergence between the distributions by making the variational posterior fit the true posterior as closely as possible, we compute the gradient of $D_{KL}$ with respect to $\phi$, obtaining:

\begin{equation}
   \nabla_{\phi}D_{\text{KL}}(q\|p) = -\frac{\partial H(q)}{\partial \phi} + \frac{\partial H(q, p)}{\partial \phi} + \cancel{\frac{\partial \ln p(s; \theta)}{\partial \phi}},
\label{eq:13}
\end{equation}
since $p(s; \theta)$ does not depend on $\phi$ and therefore its gradient is the null vector. In the case where \( s \) is a discrete random variable, the following inequality holds:

\begin{equation}
\begin{aligned}
   D_{\text{KL}}(q\|p) = -H(q) + H(q, p) + \ln p(s; \theta) &\leq -H(q) + H(q, p)\\
   &\leq \mathcal{F},
\end{aligned}
\label{eq:14}
\end{equation}
where $\mathcal{F}= H(q, p)-H(q)$ is called \textit{variational free energy}, or simply free energy (FE).
Since the $D_{KL}$ is a non-negative quantity, and $\ln p(s; \theta) \leq 0 \,\,\, \forall s \in \mathcal{S} = \{s_1, \dots, s_n\}$, it follows that the $D_{KL}$ is always upper-bounded by the FE.  
This establishes a tractable bound on an otherwise intractable quantity, thereby allowing an intractable inference problem to be addressed as a tractable optimization one. This relationship is illustrated in the bottom-left pane of Figure~\ref{fig:dkl}. However, it is important to note that when $s$ is continuous, this property no longer holds. This is because density functions can take values greater than one, resulting in positive logarithm values. However, \eqref{eq:13} holds regardless when dealing with continuous random variables, still allowing us to optimize the intractable distance measure.

Another property of FE is that, when its sign is negated, it provides a bound on the model's evidence. The negative free energy (NFE) is therefore also known as the evidence lower bound (ELBO), a common terminology in the field of ML and statistics. This can be shown by evaluating the evidence logarithm
\begin{equation}
   \ln p(s; \theta) = \ln\left(\int_{\mathcal{X}} \frac{q(x; \phi)}{q(x; \phi)} p(s, x; \theta)\,dx\right),
\label{eq:15}
\end{equation}
where we have multiplied and divided by the same quantity $q(x; \phi)$. By applying Jensen's inequality, we can write
\begin{equation}
\begin{aligned}
   \ln\left(p(s; \theta)\right) &\geq \int_{\mathcal{X}} q(x; \phi) \ln\left(\frac{p(s, x; \theta)}{q(x; \phi)} \right)\,dx\\
   &\geq \int_{\mathcal{X}} q(x; \phi) \ln p(s, x; \theta)\,dx  -  \int_{\mathcal{X}} q(x; \phi)\ln q(x; \phi) \,dx\\
   &\geq \mathcal{\Tilde{F}},
\label{eq:16}
\end{aligned}
\end{equation}
where $\mathcal{\Tilde{F}}$ is the NFE. As shown in Figure~\ref{fig:dkl}, we can now appreciate how the (N)FE simultaneously bounds two essential but intractable quantities, \( D_{\mathrm{KL}} \) and the model log evidence, thereby recasting an intractable inference problem into a tractable optimization problem. Indeed, optimizing the (N)FE with respect to the variational posterior parameters yields an approximation of the true posterior, which can then be used as a proxy to update the model parameters and reduce surprise on the observation. This alternating procedure of optimizing the variational posterior and updating the generative parameters corresponds to the Expectation-Maximization (EM) algorithm: in the first phase, the approximate posterior over the latent variables is refined for fixed generative parameters; in the second phase, the generative parameters are updated based on the current posterior estimate. Repeatedly alternating these phases iteratively improves the estimates until convergence is reached.

\subsection*{Predictive Coding Network}
With the necessary mathematical background introduced, we can now proceed to develop a PC network. This section is divided into three subsections. The first one covers the model specification, fitting in the precedent formulas the actual models that characterize PC. The two subsequent subsections will focus on deriving the update rules for the variational and generative parameters, respectively.

\subsubsection*{Model specification}
In this subsection, we replace the earlier abstract framework with concrete functional forms to fully specify the model. Specifically, we define a Gaussian generative model and approximate the posterior distribution using a Dirac delta variational posterior. All modelling assumptions are summarized in Box \ref{box:1}. We conclude with the resulting expression for the NFE under these assumptions.

\begin{assumptionbox}[label={box:1}]{Summary of the Main Assumptions in Predictive Coding}
\begin{itemize}
    \item \textbf{Gaussian Generative Model:} Neurons are assumed to be continuous random variables distributed according to a Gaussian density.
    
    \item \textbf{Hierarchical Generative Structure:} The mean of the Gaussian densities at each level is a function parametrized by neurons in the layer above in the hierarchy and by the synapses connecting them.
    
    \item \textbf{Gaussian Variational Posterior:} The variational posterior distribution for each neuron is assumed to be Gaussian. In this work, we consider the limiting case where it reduces to a Dirac delta distribution.
    
    \item \textbf{Mean Field Approximation:} All the variational factors are assumed to be independent.
\end{itemize}
\end{assumptionbox}

To account for the hierarchical structure of the visual cortex and the complexity of real-world stimuli, we need to extend the generative model introduced in the numerator of \eqref{eq:7} to a hierarchical formulation with high-dimensional latent variables and stimuli.
We define \( x_\ell \in \mathbb{R}^{d_{\ell}} \) as the \( d_\ell \)-dimensional vector of latent variables representing the brain's neural activity encoding the most likely causes of sensory stimuli at layer \( \ell \), with \(s \in \mathbb{R}^{k} \) denoting the input sensed at the bottom layer of the hierarchy. The conditional distributions are now parametrized by parameter matrices \( \Theta_\ell \in \mathbb{R}^{d_{\ell-1} \times d_{\ell}} \), representing the synaptic connections between neurons and providing the model with greater flexibility to learn and refine the relationships between latent causes and sensory observations. Thus, the generative model is defined as
\begin{equation}
    p(x_{L:0}; \Theta_{L:1}) = p(x_L) \prod_{\ell=0}^{L-1} p(x_\ell \mid x_{\ell+1}; \Theta_{\ell+1}),
    \label{eq:17}
\end{equation}
where the latent variables at each layer are conditionally dependent on those from the layer above, with \( x_L \) representing the prior belief at the top, and \( x_0 = s \) denoting the sensory stimulus at the bottom of the hierarchy. PC assumes Gaussian latent variables, resulting in the following generative model:

\begin{equation}
    p(x_{L:0}; \Theta_{L:1}) = \mathcal{N}(x_L; \bar{\mu}, \bar{\Sigma})  \prod_{\ell=0}^{L-1} \mathcal{N}(x_{\ell}; \mu_{\ell}, \Sigma_{\ell}).
    \label{eq:18}
\end{equation}

Each Gaussian's mean is given by a parametrized function \( \mu_\ell = f_{\Theta_{\ell+1}}(x_{\ell+1}) \), with \( f_{\Theta_{\ell+1}} : \mathbb{R}^{d_{\ell+1}} \rightarrow \mathbb{R}^{d_\ell} \) representing the forward mapping from layer \( \ell+1 \) to layer \( \ell \). Note that the generative model is defined as a forward model with indices decreasing from top to bottom layers, following the indexing convention inspired by the visual cortex in PC literature \cite{rao1999predictive,friston2005theory,friston2009predictive}, which contrasts the standard increasing layer indexing used in conventional ANNs literature.
 Successively, we define the variational posterior $q(x_{L:0}; \phi_{L:0})$, which we assume to be composed of variational factors that are independent from each other, taking the form:

\begin{equation}
    q(x_{L:0}; \phi_{L:0}) = \prod_{\ell=0}^{L} q(x_\ell; \phi_\ell).
    \label{eq:19}
\end{equation}
Each variational factor is parametrized by some parameters $\phi_\ell \in \mathbb{R}^{d_\ell}$, providing the necessary degrees of freedom to better approximate the true posterior. As we will see later, depending on the settings, the latent variables in the first and/or last layers are fixed to specified values (inputs and targets) and therefore do not require inference.

The variational posterior is often modelled as a Gaussian distribution in the FEP and PC literature. While the resulting maths is more involved \cite{friston2009predictive}, a limiting case using a Dirac delta variational posterior simplifies the derivation and yields nearly identical update rules in practice. The Dirac delta function can be seen as the limit of a Gaussian as its variance approaches zero:

\begin{equation}
    \delta(x-\mu) = \lim_{\sigma \to 0} \frac{1}{\sqrt{2\pi}\sigma} e^{-\frac{(x-\mu)^2}{2\sigma^2}}.
    \label{eq:20}
\end{equation}
The motivation behind this choice is that we are mainly interested in the modes of the posterior distribution, i.e., the most likely values for the causes given the stimuli. Using a Gaussian variational posterior requires a Taylor's expansion around the mode of the log-joint $\ln p(s, x; \Theta)$ in the NFE's cross entropy term, making the assumption that the log-joint is tightly peaked around the mode. 
Using the Dirac delta, we assume no variance in the variational posterior, resulting in a deterministic approximation of the log-joint modes. This allows us to avoid Taylor expansion since the expected value of a function under a Dirac distribution centred at \(\mu\) equals the function evaluated at that concentration point:

\begin{equation}
    \mathbb{E}_{\delta}[f(x)] = \int_{\chi} \delta(x - \mu)f(x) \, dx = f(\mu).
    \label{eq:21}
\end{equation}
This holds in the multivariate case $x , \mu \in \mathbb{R}^d$ as well:
\begin{equation}
    \mathbb{E}_{\delta}[f(x)] = \int_{\chi} \prod_{i=1}^d \delta(x_i - \mu_i) f(x) \, dx = f(\mu)
    \label{eq:22}
\end{equation}
where again the multiple integral has not been expanded for brevity of notation. We consider a Dirac delta distribution parametrized by \(\phi \in \mathbb{R}^{d_\ell}\), with its density centred at the modes of the underlying generative model. This can be expressed as \(\delta(x_\ell; \phi_\ell)\), where \(\phi_\ell = \mu_\ell\). Consequently, the joint distribution is defined as
\begin{equation}
    q(x_{L:0}; \phi_{L:0}) = \prod_{\ell=0}^{L} \delta(x_\ell - \phi_\ell).
    \label{eq:23}
\end{equation}
Before fitting our models into the NFE, we must address the entropy term. In the discrete case, the Dirac delta's entropy is well-behaved and equals zero, but in the continuous case, the differential entropy tends to $-\infty$. However, it can be considered as a constant and thus still disregarded in the optimization process. Considering only the cross entropy term, we then have

\begin{equation}
\begin{aligned}
     \mathcal{\Tilde{F}} &= - H(q, p) + \cancel{H(q)}\\
     &= \int_{\mathcal{X}}\prod_{\ell=0}^{L} q(x_\ell; \phi_\ell)\ln\left(p(x_L) \prod_{\ell=0}^{L-1} p(x_\ell|x_{\ell+1}; \Theta_{\ell+1})\right) \, dx\\
     &= \int_{\mathcal{X}}\prod_{\ell=0}^{L} \delta(x_\ell - \phi_\ell)\ln\left(\mathcal{N}(x_L; \bar{\mu}, \bar{\Sigma})  \prod_{\ell=0}^{L-1} \mathcal{N}(x_{\ell}; \mu_{\ell}, \Sigma_{\ell})\right) \, dx,\\
\end{aligned}
    \label{eq:24}
\end{equation}
and by applying the logarithms sum property we obtain

\begin{equation}
\begin{aligned}
     \mathcal{\Tilde{F}} &= \int_{\mathcal{X}}\prod_{\ell=0}^{L} \delta(x_\ell - \phi_\ell) \left[ \ln\left(\mathcal{N}(x_L; \bar{\mu}, \bar{\Sigma})\right) +  \sum_{\ell=0}^{L-1} \ln\left(\mathcal{N}(x_{\ell}; \mu_{\ell}, \Sigma_{\ell})\right) \right] \, dx.
\end{aligned}
    \label{eq:25}
\end{equation}
We can now apply the Dirac delta's property described in \eqref{eq:22}, obtaining:

\begin{equation}
     \mathcal{\Tilde{F}} = \sum_{\ell=0}^{L} \ln\left(\mathcal{N}(\phi_{\ell}; \mu_{\ell}, \Sigma_{\ell})\right),
\label{eq:26}
\end{equation}
resulting in a sum of Gaussians evaluated at the Dirac's center of mass, where $\mu_L=\bar{\mu}$ and $\Sigma_L={\bar{\Sigma}}$. Proceeding with expanding the NFE, we have 

\begin{equation}
\begin{aligned}
     \mathcal{\Tilde{F}} &= \sum_{\ell=0}^{L} - \frac{d_\ell}{2} \ln(2\pi) -\frac{1}{2} \ln|\Sigma_\ell| -\frac{1}{2} (\phi_\ell - \mu_\ell)^T \Sigma_{\ell}^{-1} (\phi_\ell - \mu_\ell)\\
     &\approx -\frac{1}{2} \left[ \sum_{\ell=0}^{L}   \ln|\Sigma_\ell| + (\phi_\ell - \mu_\ell)^T \Sigma_\ell^{-1} (\phi_\ell - \mu_\ell) \right],
\end{aligned}
    \label{eq:27}
\end{equation}
where the constants have been discarded, as they do not depend on the parameters and thus do not affect the optimization process. Crucially, the objective function consists only of layer-wise precision-weighted error terms between the variational posterior values and the generative model predictions, expressed as \( \mu_\ell = f_{\Theta_{\ell+1}}(\phi_{\ell+1}) \), following from~\eqref{eq:22}.

\subsubsection*{Variational parameters optimization}
In this subsection, we focus on optimizing the NFE with respect to the variational parameters to approximate (the modes of) the true posterior distribution. We also show how this optimization results in local update rules, enabling the neural activity to be updated using only information from adjacent layers.

\begin{figure}[h!]
    \centering
    \includegraphics[width=0.99\textwidth]{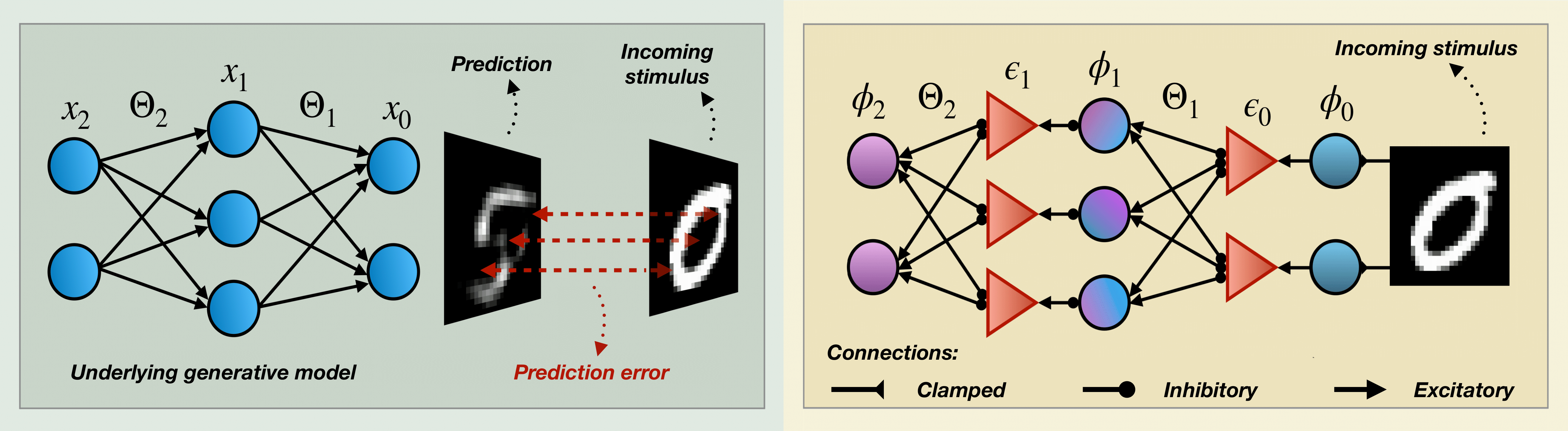}
    \caption{\doublespacing The left pane illustrates PC’s probabilistic graphical model, showing hierarchical dependencies between variables. The model’s prediction, based on the most likely value for each variable, corresponds to a standard neural network forward pass. At the hierarchy’s bottom, sensed stimuli are compared to predictions, generating error signals. During neural activity optimization, the lowest layer is clamped to the stimuli, while the other layers update to match these stimuli according to the generative model’s forward pass, as illustrated in the right pane. Red triangles denote error nodes, which receive signals from circle-shaped value neurons from the layer below and from the layer above. The figure assumes the covariance matrix is fixed to the identity and is therefore omitted.}
    \label{fig:pcnet2}
\end{figure}

PC distinguishes between two types of neurons: error neurons and value neurons. The former, represented as red triangles in the right pane of Figure~\ref{fig:pcnet2}, encode the precision-weighted prediction errors at each layer. The latter, represented as coloured circles, encode the modes of the approximated posterior distribution.
The network shown in Figure~\ref{fig:pcnet2} illustrates the generative setting, where the input stimuli are clamped to the value neurons at the lowest layer, which typically corresponds to the output layer of a neural network. The highest layer, which usually corresponds to the input layer, is left free to converge to the most likely configuration whose forward pass would produce the clamped stimuli given the current network parameters.

\begin{figure}[h]
    \centering
    \includegraphics[width=0.99\textwidth]{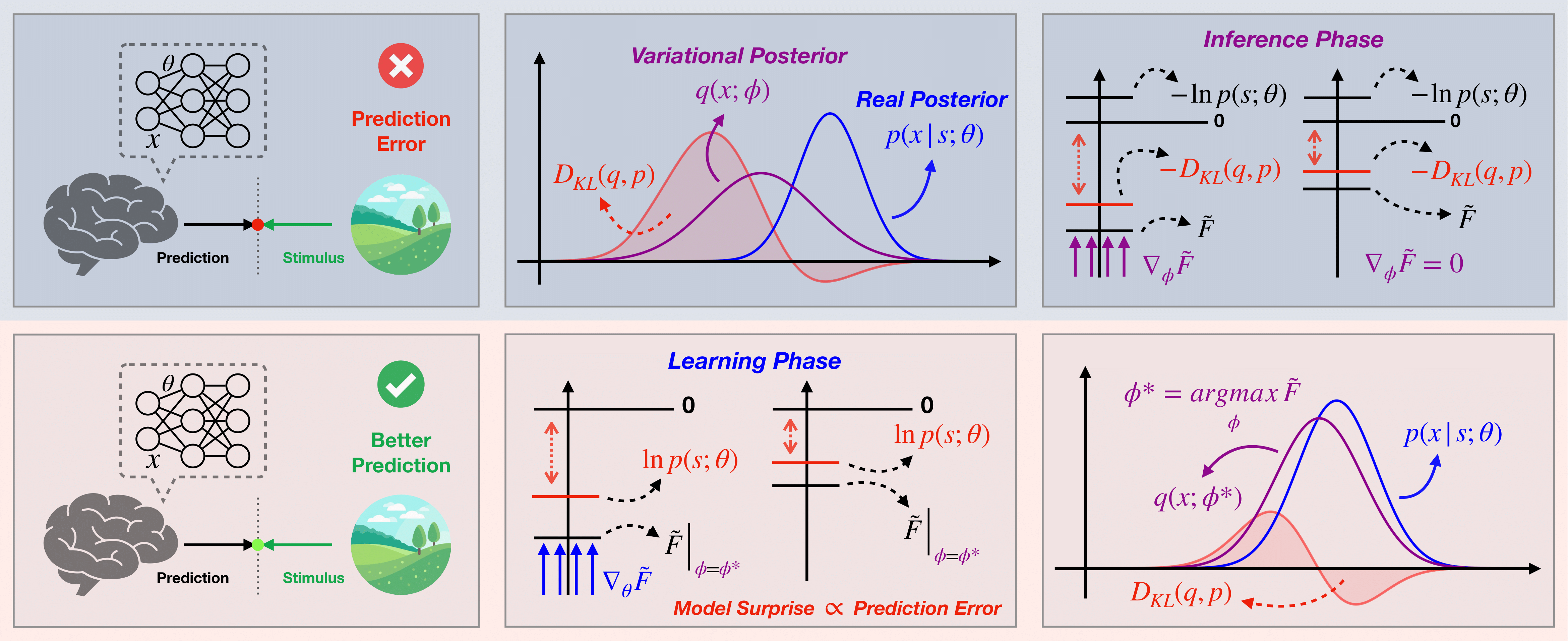}
    \caption{\doublespacing The figures summarize the steps involved in PC. The upper-left pane shows a prediction error when the perceived signal differs from the brain’s prediction. Due to the intractability of the true posterior, a variational posterior is used for approximation, with the mismatch quantified by the Kullback-Leibler divergence (red area in the upper-center pane). The upper-right pane illustrates how optimizing a tractable lower bound improves the variational posterior. The lower-right pane shows the improved posterior approximation computed during the inference phase, and the lower-center pane illustrates how this approximation is used to adjust generative model parameters, optimizing the network prediction\color{black}}
    \label{fig:pcnutshell}
\end{figure}

Building on this framework, we define the error nodes at a layer \(\ell\) as \(\epsilon_\ell \triangleq \Sigma_\ell^{-1} (\phi_\ell - \mu_\ell)\), and the layer-wise energy as \(\xi_\ell \triangleq (\phi_\ell - \mu_\ell)^T \Sigma_\ell^{-1} (\phi_\ell - \mu_\ell)\). Using these definitions, we can express the gradient of the NFE with respect to \(\phi_\ell\) as follows:

\begin{equation}
\begin{aligned}
     \frac{\partial \tilde{\mathcal{F}}}{\partial \phi_\ell} &= -\frac{1}{2} \left[ \frac{\partial}{\partial \phi_\ell} \sum_{\ell=0}^{L}   \ln|\Sigma_\ell| + (\phi_\ell - \mu_\ell)^T \Sigma_\ell^{-1} (\phi_\ell - \mu_\ell) \right] \\ 
    &= -\frac{1}{2} \left( \frac{\partial \xi_\ell}{\partial \phi_\ell}  + \frac{\partial \xi_{\ell-1}}{\partial \phi_\ell}\right),
\end{aligned}
\label{eq:28}
\end{equation}
as \(\phi_{\ell}\) affects only layers \(\ell\) and \(\ell-1\). We proceed by computing the gradient of the two energy terms separately, starting from the first one:
\begin{equation}
\begin{aligned}
\frac{\partial \xi_\ell }{\partial \phi_\ell} &= \frac{\partial}{\partial \phi_\ell}(\phi_\ell - f_{\Theta_{\ell+1}}( \phi_{\ell+1}))^T \Sigma_\ell^{-1} (\phi_\ell - f_{\Theta_{\ell+1}}(\phi_{\ell+1})).
\end{aligned}
\label{eq:29}
\end{equation} 
Since the inverse covariance matrix is symmetric, we can apply the following quadratic form derivative rule from matrix calculus:

\begin{equation}
\frac{\partial}{\partial x}(x - s)^T \, W(x - s) = 2W(x - s),
\label{eq:30}
\end{equation}
resulting in

\begin{equation}
\frac{\partial \xi_\ell }{\partial \phi_\ell} = 2\Sigma_\ell^{-1}(\phi_\ell - f_{\Theta_{\ell+1}}(\phi_{\ell+1})) = 2 \epsilon_\ell,
\label{eq:31}
\end{equation}
where we have applied the error node definition.
To compute the gradient of the second energy term, we use another derivative rule for the quadratic form:
\begin{equation}
\frac{\partial}{\partial s}(x - s)^T \, W(x - s) = -2W(x - s).
\label{eq:32}
\end{equation}
By applying it, we obtain
\begin{equation}
\begin{aligned}
\frac{\partial \xi_{\ell-1}}{\partial \phi_\ell}  &= \frac{\partial}{\partial \phi_\ell}(\phi_{\ell-1} - f_{\Theta_{\ell}}(\phi_{\ell}))^T \Sigma_{\ell-1}^{-1} (\phi_{\ell-1} - f_{\Theta_{\ell}}(\phi_{\ell}))\\
 &= -2 \left(\frac{\partial }{\partial \phi_\ell}f_{\Theta_{\ell}}(\phi_{\ell})\right)^T \epsilon_{\ell-1}.
\end{aligned}
\label{eq:33}
\end{equation}
Now we can substitute \eqref{eq:31} and \eqref{eq:33} inside \eqref{eq:28}, obtaining:
\begin{equation}
\frac{\partial \tilde{\mathcal{F}}}{\partial \phi_\ell} 
= \left(\frac{\partial }{\partial \phi_\ell}f_{\Theta_{\ell}}(\phi_{\ell})\right)^T \epsilon_{\ell-1} - \epsilon_\ell,
\label{eq:34}
\end{equation}
where $f_{\Theta_{\ell}}(\phi_{\ell}): \mathbb{R}^{d_\ell} \rightarrow \mathbb{R}^{d_{\ell-1}}$, and the Jacobian of $f$, denoted by $\frac{\partial }{\partial \phi_\ell}f_{\Theta_\ell}(\phi_\ell)$, belongs to $\mathbb{R}^{d_{\ell-1} \times d_{\ell}}$, consistently with $\epsilon_{\ell-1} \in \mathbb{R}^{d_{\ell-1}}$ and $\epsilon_\ell \in \mathbb{R}^{d_{\ell}}$. During the NFE optimization with respect to the variational parameters, called the \textit{inference phase}, the lowest and/or highest layers are typically clamped to the stimuli, the observation, or both, depending on the task. For example, in the generative setting illustrated in Figure \ref{fig:pcnet2}, only the lowest layers are clamped. In the classification setting, both the highest and lowest layers are clamped to the input stimuli and output labels, respectively. Consequently, their values are known and do not require inference. The variational parameters can be modelled either as a discrete-time system with $\Delta t = 1$, or as a continuous-time dynamical system, i.e., $\frac{d\phi_\ell}{dt} \propto \frac{\partial \tilde{\mathcal{F}}}{\partial \phi_\ell}$, resulting in a set of differential equations that can be solved using an Euler scheme:

\begin{equation}
    \phi_\ell(t + \Delta t) = \phi_\ell(t) + \alpha \, \Delta t \, \frac{\partial \tilde{\mathcal{F}}}{\partial \phi_\ell},
    \label{eq:35}
\end{equation}
where $\Delta t$ defines the time resolution, and $\alpha$ is the learning rate, which scales the step size (the latter can be incorporated in the former, and vice versa). Now, by looking at the right pane of Figure \ref{fig:pcnet2} and \eqref{eq:34}, we can understand how the dynamics of the variational parameters depend on inhibitory feedforward signals from the error nodes above and from excitatory feedback signals from the error nodes below, where the latter carry information about the wrong information in the prediction. This optimization should be interpreted as climbing the dashed black function in the bottom-left pane of Figure \ref{fig:dkl}, and is run until convergence, where we have

\begin{equation}
    \phi^* = \underset{\phi}{\argmax} \, \tilde{\mathcal{F}}.
    \label{eq:36}
\end{equation}
The process of optimizing the NFE with respect to neural activity is illustrated in the top row of Figure~\ref{fig:pcnutshell}, with the final approximation shown in the bottom-right pane. Note that a Gaussian variational posterior is used there to illustrate the general PC scenario. When a Dirac delta is used instead, the purple distribution collapses to a point mass, and the NFE is maximized as this point converges to the mode of the true posterior.

\subsubsection*{Generative parameters optimization}
In this subsection, we focus on optimizing the NFE with respect to the generative parameters, using the variational posterior approximation as a proxy for the true posterior. We will also demonstrate how these updates are biologically plausible by relying solely on locally available information.

The process of updating the generative parameters to improve the network’s predictions is referred to as the \textit{learning phase}. It requires evaluating the gradient of the NFE with respect to $\Theta_\ell$, for which we adopt the notation $f_{\phi_\ell}(\Theta_\ell)$ instead of $f_{\Theta_\ell}(\phi_\ell)$ to emphasize that $\Theta$ is the variable of interest. This results in:

\begin{equation}
\begin{aligned}
\frac{\partial \tilde{\mathcal{F}}}{\partial \Theta_\ell} &= -\frac{1}{2} \left[ \frac{\partial}{\partial \Theta_\ell} \sum_{\ell=0}^{L}   \ln|\Sigma_\ell| + \xi_{\ell} \right] \\ 
 &= -\frac{1}{2} \left[ \frac{\partial}{\partial \Theta_{\ell}}(\phi_{\ell-1} - f_{\phi_\ell}(\Theta_\ell))^T \Sigma_{\ell-1}^{-1} (\phi_{\ell-1} - f_{\phi_\ell}(\Theta_\ell))\right] .
\end{aligned}
\label{eq:37}
\end{equation}
As $\Sigma^{-1}$ is symmetric, we can apply the chain rule of calculus and \eqref{eq:32}, obtaining: 

\begin{equation}
\begin{aligned}
\frac{\partial \tilde{\mathcal{F}}}{\partial \Theta_\ell}&= -\frac{1}{2} \left[ -2 \Sigma_{\ell-1}^{-1}(\phi_{\ell-1} - f_{\phi_\ell}(\Theta_\ell)) \left( \frac{\partial}{\partial \Theta_\ell} f_{\phi_\ell}(\Theta_\ell) \right) \right] \\
&= f'_{\phi_\ell}(\Theta_\ell) \odot \epsilon_{\ell-1} \phi_\ell^T,
\end{aligned}
\label{eq:38}
\end{equation}
assuming \(f\) is a linear transformation followed by a nonlinearity, i.e., 
\( f_{\phi_\ell}(\Theta_\ell) = f(\Theta_\ell \phi_\ell) \).  
As with the variational parameters, we have \( f_{\phi_\ell}(\Theta_\ell): \mathbb{R}^{d_\ell} \to \mathbb{R}^{d_{\ell-1}} \), with \( f'_{\phi_\ell}(\Theta_\ell), \epsilon_{\ell-1} \in \mathbb{R}^{d_{\ell-1}} \) and \( \phi_\ell^\top \in \mathbb{R}^{1 \times d_\ell} \).

To optimize \eqref{eq:27}, the generative parameters can be updated proportionally to this derivative, using the optimal values $\phi^*$ found for the variational parameters as a proxy for the real maximum a posteriori value. PC alternates between updating the variational parameters to approximate the true posterior given the current generative parameters, and then using these variational parameters as a proxy for the posterior to optimize the generative parameters. This exactly follows the EM algorithm described at the end of the VI section of this manuscript. However, unlike the standard EM algorithm, only a small gradient step $\Delta\Theta_\ell$ is taken during generative parameters optimization, to avoid overfitting a specific set of stimuli:

\begin{equation}
    \Theta_\ell(t + \Delta t) = \Theta_\ell(t) + \eta \frac{\partial}{\partial \Theta_\ell}\tilde{\mathcal{F}}\bigg|_{\phi=\phi^*},
    \label{eq:39}
\end{equation}

where $\eta$ is the learning rate of the generative parameters. As a result, unexpected but rare inputs caused by noise have a weak impact on the generative parameters, while consistent statistical regularities produce stronger parameter updates. As for the neural activity, alternatively to a discrete gradient step, the generative parameters can also be formulated as a continuous-time system. In this case, the generative parameters evolve over a much shorter period of time according to $\frac{d\Theta_\ell}{dt} \propto \frac{\partial}{\partial \Theta_\ell}\tilde{\mathcal{F}}\bigg|_{\phi=\phi^*}$. This perspective enforces a separation of timescales, where the variational parameters can be interpreted as rapid neural activity changes, while the generative parameters undergo slower synaptic adjustments. Limiting convergence to a short period is crucial to prevent overfitting, similar to the role of a small learning rate in the discrete case.

The generative model's covariance matrices $\Sigma_\ell$ enable PC networks to modulate gradients based on the associated uncertainty. Although this modulation of errors can be seen as an advantage of PC over BP, it remains a relatively unexplored research topic. Furthermore, learning the covariance matrices involves two challenges arising from the necessity of computing their inverse during gradient evaluation.
First, computing the inverse requires a neuron to access information from neurons to which it is not directly connected, such as the variance of other units, which violates the principle of local computation. Second, if some components of the covariance matrix tend to zero, their inverse diverges, potentially causing the gradient to explode and leading to numerical instability during training. For instance, in the case of a diagonal covariance matrix, the inverse has entries $\frac{1}{\sigma^2}$ on the diagonal; therefore, if neurons exhibit variance approaching zero, the inverse will diverge, as the gradients computed in \eqref{eq:34} and \eqref{eq:38}.

Nevertheless, for the sake of completeness, we include the derivation of a possible learning rule for $\Sigma_\ell$ based on the NFE gradient, analogous to those for $\phi_\ell$ and $\Theta_\ell$ discussed earlier:
\begin{equation}
\begin{aligned}
\frac{d \Sigma_\ell}{dt} \propto \frac{\partial \tilde{\mathcal{F}}}{\partial \Sigma{_\ell}} &=  -\frac{1}{2} \frac{\partial}{\partial \Sigma{_\ell}} \sum_{\ell=1}^{L} (\phi_\ell - \mu_\ell)^T \Sigma_\ell^{-1} (\phi_\ell - \mu_\ell) + \ln|\Sigma_\ell| \\
&= -\frac{1}{2} \left[ \frac{\partial}{\partial \Sigma{_\ell}} (\phi_\ell - \mu_\ell)^T \Sigma_\ell^{-1} (\phi_\ell - \mu_\ell) + \Sigma_\ell^{-1} \right].
\end{aligned}
\label{eq:40}
\end{equation}
Given the following matrix derivative rule, which assumes that \( A \) is a symmetric, invertible, and square matrix, such that the identity \( A^{-1} = (A^{-1})^T \) holds:
\begin{equation}
\frac{\partial}{\partial A} \left( x^T A^{-1} x \right) = - A^{-1} x x^T A^{-1} = - (A^{-1} x)(A^{-1} x)^T,
\label{eq:41}
\end{equation}
we can write:
\begin{equation}
\begin{aligned}
\frac{\partial \tilde{\mathcal{F}}}{\partial \Sigma{_l}} &= -\frac{1}{2} \left[ -(\Sigma_l^{-1} (\phi_l - \mu_l))(\Sigma_l^{-1} (\phi_l - \mu_l))^T + \Sigma_l^{-1} \right] \\
&= \frac{1}{2} \left[\epsilon_l \epsilon_l^T - \Sigma_l^{-1} \right].
\end{aligned}
\label{eq:42}
\end{equation}
which would allow us to define an update rule analogous to \eqref{eq:35} and \eqref{eq:39}.

\begin{tcolorbox}[
    colback=blue!5!white,
    colframe=blue!75!black,
    colbacktitle=purple!20!blue!75,
    coltitle=white,
    title=Box 2 - Summary of Predictive Coding Updating Rules,
    fonttitle=\bfseries,
    enhanced,
    frame hidden,
    drop fuzzy shadow
]
\textbf{1. Update rule for Variational Parameters:}  
\[
\phi_\ell(t + \Delta t) = \phi_\ell(t) 
+ \Delta t \big( f'(\Theta_\ell\phi_\ell) \odot \Theta^T \epsilon_{\ell-1} - \epsilon_\ell \big)
\]

\textbf{2. Update rule for Generative Parameters:}  
\[
\Theta_\ell(t + \Delta t) = \Theta_\ell(t) 
+ \Delta t \big( f'(\Theta_\ell\phi_\ell) \odot \epsilon_{\ell-1} \phi_\ell^T \big)
\]
\[
\Sigma_\ell(t + \Delta t) = \Sigma_\ell(t) 
+ \frac{\Delta t}{2} \big( \epsilon_\ell \epsilon_\ell^T - \Sigma_\ell^{-1} \big)
\]
\end{tcolorbox}

\subsection*{Predictive Coding and Backpropagation}
In this section, we analyze and compare the key differences between PC and the BP algorithm, following the approach of \cite{whittington2017approximation}, and refer to other relevant studies for interested readers.

Let us assume we have an ANN whose loss function $\mathcal{L}$ is the least squares function

\begin{equation}
\mathcal{L} = \frac{1}{2} \| y_L - \hat{y} \|_2^2,
\label{eq:45}
\end{equation}
where $\hat{y}$ denotes the target vector and $y_L$ the output vector of the network. The objective is to optimize this loss function by iteratively updating the ANN's parameters in order to obtain predictions that better match the desired targets. The ANN forward pass is modelled as linear transformation followed by a nonlinearity:

\begin{equation}
y_{\ell+1} = f(W_{\ell} y_{\ell})
\label{eq:46}
\end{equation}
for \(0 \leq \ell < L\), where \(f(\cdot)\) represents any differentiable and continuous activation function, and \(W_{\ell-1}\) denotes the parameter matrix of layer \(\ell-1\). The parameter updates are performed by evaluating the loss gradient using the chain rule of calculus: 

\begin{equation}
\frac{\partial \mathcal{L}}{\partial W_{\ell}} = \frac{\partial \mathcal{L}}{\partial y_L} \left( \prod_{k=L}^{\ell+2} \frac{\partial y_k}{\partial z_{k}} \frac{\partial z_{k}}{\partial y_{k-1}} \right) \frac{\partial y_{\ell+1}}{\partial z_{\ell+1}} \frac{\partial z_{\ell+1}}{\partial W_\ell},
\label{eq:47}
\end{equation}
where  $z_{\ell+1} = W_{\ell} y_{\ell}$. This expression can be reformulated recursively by introducing the error terms \(\delta^\ell\), defined as:

\begin{equation}
\delta_{\ell} = \begin{cases} 
(y_\ell - \hat{y}) & ,\, \ell = L \\
 f'(W_{\ell} y_{\ell}) \odot W_{\ell}^T\delta_{\ell+1} & ,\, 0 < \ell < L,
\end{cases}
\label{eq:48}
\end{equation}
resulting in the following gradient:
\begin{equation}
\frac{\partial \mathcal{L}}{\partial W_\ell} = \delta_{\ell+1} y_\ell^T.
\label{eq:49}
\end{equation}
Interestingly, a recursive formulation can also be derived in a particular case of PC. Such recursion arises by setting the gradient in \eqref{eq:34} to zero, meaning the NFE is maximized with respect to the variational parameters, and the network's activity has reached equilibrium:

\begin{equation}
\begin{aligned}
\frac{\partial \tilde{\mathcal{F}}}{\partial \phi_\ell} &= 0 \\
\epsilon_{\ell} &= f'(\Theta_\ell\phi_\ell) \odot \Theta^T \epsilon_{\ell-1}.
\label{eq:50}
\end{aligned}
\end{equation}
To simplify the comparison between the two optimization algorithms, we invert the indexing of the PC model, treating the highest layer as the $0$-th and the lowest as the $L$-th, in contrast to Figure~\ref{fig:pcnet2}. For a more direct comparison, we consider PC in the supervised learning setting. In this context, the inputs are clamped to the $0$-th layer, while the labels are clamped to the $L$-th layer. The PC error equation can now be written recursively as follows:

\begin{equation}
\epsilon_{\ell} = \begin{cases} 
\Sigma_{L}^{-1}(\hat{y} - \mu_{L}) & ,\, \ell = L \\
 f'(\Theta_\ell\phi_\ell) \odot \Theta_{\ell}^{T} \epsilon_{\ell+1} & , \, 0 < \ell < L,
\end{cases}
\label{eq:51}
\end{equation}
as in the last layer of the hierarchy, the error is the precision-weighted difference between the model's prediction and the clamped target vector $\hat{y}$. Comparing \eqref{eq:48} and \eqref{eq:51}, we can see an explicit similarity, allowing us to show how PC can approximate BP gradients, assuming the two models to be initialized with the same generative parameters, and the covariance matrix $\Sigma_L$ fixed to the identity matrix. 
After both networks predicts the output for a given input, the PC network starts from an equilibrium state, satisfying \eqref{eq:50}. Then, its output neurons are clamped to the target values. If the initial prediction was incorrect, this clamping perturbs the error neurons at layer $L$  generating feedback error signals.
These signals cause changes in the activity of PC model's neurons according to \eqref{eq:35}, resulting in deviations from the forward-pass activity, which was initially equivalent in both BP and PC models. Consequently, even if \eqref{eq:50} is satisfied, the weight updates may deviate from those obtained via BP.
Therefore, to approximate BP gradients, we need to mitigate the change in the PC model's neural activity. To achieve this, the covariance (identity) matrix in the final layer can be scaled up by a constant factor. As the matrix is inverted, this proportionally reduces the magnitude of the final layer's error neurons $\epsilon_L$ in \eqref{eq:51}. As a result, for large rescaling constants, PC neuron activity remains nearly unchanged and closely aligns with BP, as both models share initial parameters and forward pass. Consequently, the error terms in \eqref{eq:51} approximate those of BP, scaled by the rescaling factor. Thus, while the gradients of the two models point in similar directions, their magnitudes differ. To match the weight updates, PC gradients must be rescaled by the same factor used to suppress the error signals, compensating for the magnitudes difference. For readers interested in a more detailed discussion on this property, we refer to \cite{whittington2017approximation}. Some studies have also demonstrated that PC can compute exact BP gradients with additional minor modifications, such as updating parameters only at specific timestamps \cite{song2020can} or that this property can be extended to any graph topology, and not being restricted to ANN-like structures \cite{millidge2022predictive}. For those particularly interested in the comparison between BP and PC, \cite{zahid2023predictive} provides a critical review, formulating PC as a steady-state implementation of BP. The authors argue that this reinterpretation compromises essential aspects of traditional PC, including its variational Bayesian interpretation and its capacity to represent uncertainty over latent states, and they critically assess the efficiency of such approaches.

\subsection*{Predictive Coding and Kalman Filtering}
In the previous sections, we showed how a Bayesian inference problem can be reformulated as an optimization problem, resulting in a bio-inspired algorithm with spatially local updates. In this section, we show that this locality property can also be extended temporally by introducing recurrent connections into the model.

\begin{figure}[h]
    \centering
    \includegraphics[width=0.99\textwidth]{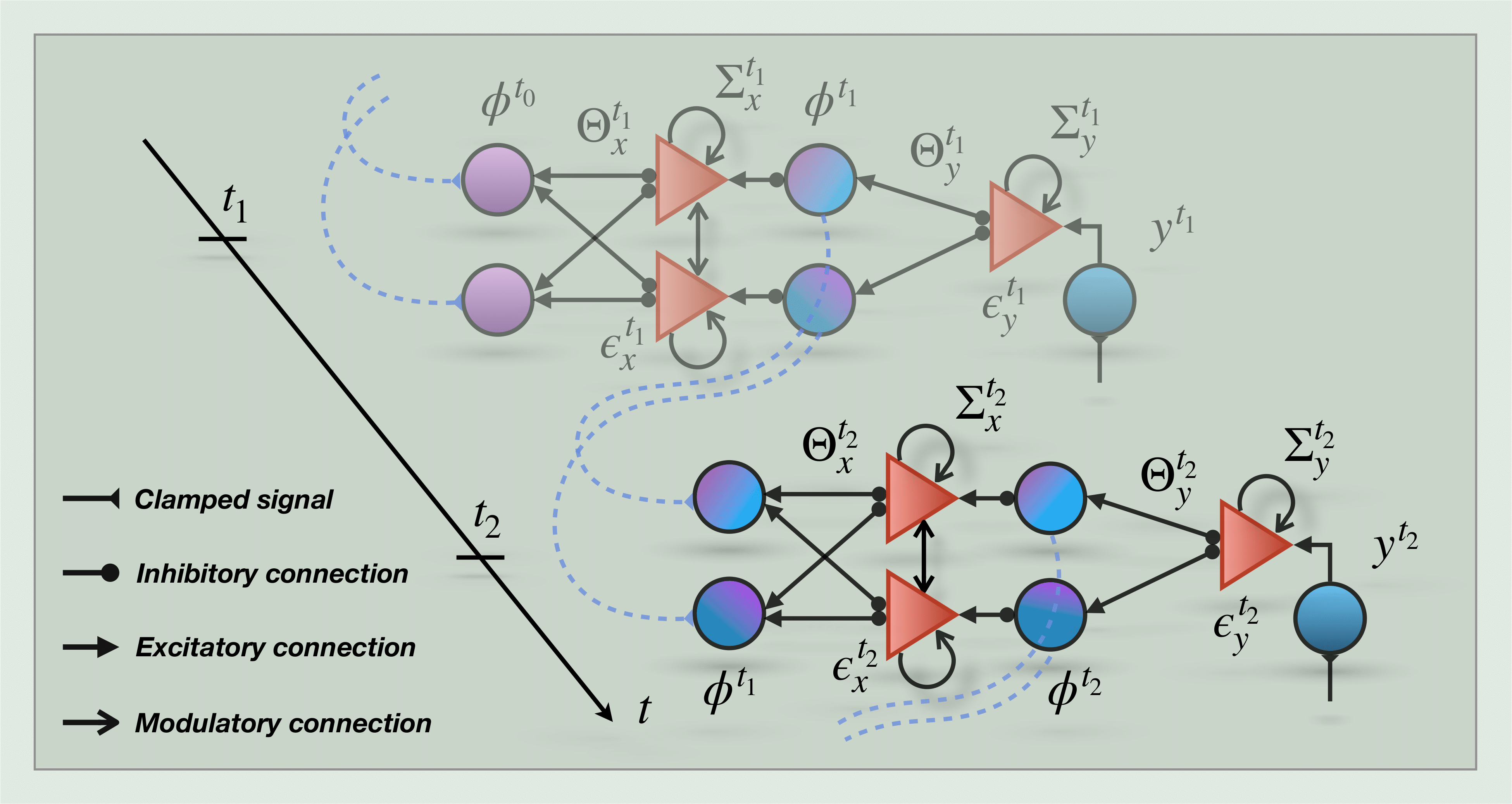}
    \caption{\doublespacing Graphical representation of the circuit derived from the PC approximation of KF. For consistency of notation, $\mu$ is renamed $\phi$, and matrices $A$ and $C$ are relabelled $\Theta_x$ and $\Theta_y$. Dashed lines indicate how each estimate serves as the prior for the next timestamp. At each timestamp, a new observation is clamped to the $y$ layer (KF measurement), and the most likely state is inferred by integrating prior and observation, repeating the process over time.\color{black}}
    \label{fig:kfpc}
\end{figure}

By augmenting a PC network with recurrent connections, it can process temporal data while maintaining local parameter updates in both space and time \cite{millidge2024predictive}. In this setting, PC becomes analogous to recursively estimating an unknown probability density function over time using incoming measurements and previous estimates. This approach is equivalent to Bayesian filtering. A particular case of the latter arises when the system is a linear Markov process and the noise is assumed to be Gaussian. In this scenario, the process reduces to the well-known KF algorithm. KF provides a closed-form solution for the Bayesian-optimal estimate of the system's next state, along with the associated uncertainty. This connection is particularly interesting, as it links neural processing to Bayesian filtering, thereby supporting the Bayesian brain hypothesis. This hypothesis proposes that the brain constructs beliefs and perceptions by integrating sensory input with prior knowledge and expectations, an assumption that aligns closely with the principles of the PC framework. We will first provide a brief overview of KF and then show how a specific configuration of recurrent PC, illustrated in Figure \ref{fig:kfpc}, results in the optimization the same objective \cite{millidge2021neural}.  Note that, as the focus of this manuscript is on PC, which concerns perception, we omit the KF's control parameter.

KF addresses the problem of trying to estimate the state $x \in \mathbb{R}^{n}$ of a linear dynamic system that is assumed to evolve accordingly to
\begin{equation}
x_{t+1} = A x_t + w\,, \quad w \sim \mathcal{N}(0, \Sigma_w),
\label{eq:50}
\end{equation}
where $A \in \mathbb{R}^{n \times n}$ is the state transition matrix of the process from state at time $t$ to time $t+1$, $x_{t} \in \mathbb{R}^{n}$ is our expected state vector at time t, and $w \in \mathbb{R}^{n}$ is a random vector representing the process noise. Therefore, we have that our estimate of the system's next state is distributed as follows:
\begin{equation}
x_{t+1} \sim \mathcal{N}(\mathbb{E}\left[x_{t+1}\right], \Sigma_{x_{t+1}}),
\label{eq:51}
\end{equation}
The Gaussian is centred at the expected projected state, which is equal to the system transition matrix multiplied by the maximum a posteriori (MAP) of the previous timestamp
\begin{equation}
\mathbb{E}\left[x_{t+1}\right] =  \mathbb{E}\left[A x_t + w\right]   = A \mu_t.
\label{eq:52}
\end{equation}
Assuming zero covariance between $x$ and $w$, the covariance matrix of the projected state value is given by
\begin{equation}
\Sigma_{x_{t+1}} = \mathrm{Cov}(A x_t + w)
= A\, \mathrm{Cov}(x_t)\, A^T + \mathrm{Cov}(w)
= A \Sigma_{x_t} A^T + \Sigma_w.
\label{eq:53}
\end{equation} 
The measurements $y \in \mathbb{R}^{m}$ performed to update our posterior belief on the state at a given time $t+1$ are modelled by 
\begin{equation}
y_{t+1} = C x_{t+1} + z\,, \quad z \sim \mathcal{N}(0, \Sigma_z),
\label{eq:54}
\end{equation}
where $C \in \mathbb{R}^{m \times n}$ is the emission matrix, mapping the state vector into the measurement vector, and $z \in \mathbb{R}^{m}$ is a random vector representing the measurement noise. Therefore, we can model the measurement vector $y_{t+1}$ as distributed according to a Gaussian distribution with mean $Cx_{t+1}$, and covariance $\Sigma_z$:
\begin{equation}
y_{t+1} \sim \mathcal{N}(Cx_{t+1}, \Sigma_{z}).
\label{eq:55}
\end{equation}

The state projection in \eqref{eq:50} acts as a prior distribution over the system, while the measurement in \eqref{eq:54} acts as a likelihood term, leading to a posterior probability distribution that combines the predicted and observed estimates:
\begin{equation} p(x_{t+1}|y_{t+1}, \mu_t) \propto p(y_{t+1}|x_{t+1}) p(x_{t+1}|\mu_t), 
\label{eq:56} 
\end{equation} where $\mu_t$ is the MAP estimate from the previous time step. \color{black} The posterior distribution is then used to predict again the evolution of the system, generating the prior belief for the subsequent timestamp. KF gives a closed-form exact solution to this problem, which is optimal in a Bayesian sense as it minimizes the mean square error (MSE) in the estimated parameters, finding the next state MAP
\begin{equation}
\mu_{t+1} = \underset{x_{t+1}}{\text{argmax}} \, \left\{p\left(x_{t+1}|y_{t+1},\mu_t\right)\right\}.
\label{eq:57}
\end{equation}
This inference problem can also be formulated from an optimization perspective \cite{millidge2021neural}. To do so, as we are mainly interested in finding the MAP of the system's state posterior distribution, we can again approximate the posterior distribution by a Dirac delta centred at the posterior's mode  $\delta(x_{t+1} - \mu_{t+1})$. This involves the same steps we already covered before, starting by lower-bounding the $D_{\text{KL}}$ between the variational posterior and the true posterior by the NFE: \color{black}

\begin{equation}
\begin{aligned}
\mathcal{\Tilde{F}} &= - H(q, p) + \cancel{H(q)}\\
&= -\mathbb{E}_{\delta} \left[ -\ln\left(\mathcal{N}(y_{t+1};Cx_{t+1},\Sigma_{z})\mathcal{N}(x_{t+1};A\mu_{t}, \Sigma_{x_{t+1}})\right) \right] \\
&= \ln\left(\mathcal{N}(y_{t+1};C\mu_{t+1},\Sigma_{z})\mathcal{N}(\mu_{t+1};A\mu_{t}, \Sigma_{x_{t+1}})\right).
\end{aligned}
\label{eq:58}
\end{equation}
Note that here $\mu$ has to be interpreted as the variational parameter, analogous to $\phi$ in the PCN section, to emphasize that as it approaches the NFE maximum, it converges to the value in \eqref{eq:57}, corresponding to the posterior modes. \color{black} Indeed, maximizing the NFE with respect to $\mu_{t+1}$ is equivalent to finding the $x_{t+1}$ value maximizing the posterior distribution in \eqref{eq:59}, as the maximum of the log-joint is the same as the maximum of the joint distribution:
\begin{equation}
\begin{aligned}
\underset{x_{t+1}}{\text{argmax}} \left\{p\left(x_{t+1}|y_{t+1}, \mu_t\right)\right\} &= \underset{x_{t+1}}{\text{argmax}} \left\{p\left(y_{t+1}|x_{t+1}) p(x_{t+1}| \mu_t\right)\right\} \\
&=\underset{x_{t+1}}{\text{argmax}} \left\{ \ln\left(\mathcal{N}(y_{t+1};Cx_{t+1},\Sigma_{z})\mathcal{N}(x_{t+1};A\mu_{t}, \Sigma_{x_{t+1}})\right) \right\}.
\end{aligned}
\label{eq:59}
\end{equation}
By expanding the two Gaussian distributions and then applying the properties of logarithms, we obtain
\begin{equation}
\tilde{\mathcal{F}} 
\approx  -\frac{1}{2} \left[ (y_{t+1} - C\mu_{t+1})^T \Sigma_{z}^{-1} (y_{t+1} - C\mu_{t+1}) + (\mu_{t+1} - A\mu_{t})^T \Sigma_{x_{t+1}}^{-1} (\mu_{t+1} - A\mu_{t}) \right],
\label{eq:60}
\end{equation}
where the normalization constants have not been considered, as they do not affect the optimization with respect to the modes. We can now use gradient ascent to iteratively approximate the most likely state of the system, after performing the prediction and measurement steps. The update rules derived from this approach are again biologically plausible updating rules that could be implemented within a neural circuit such as the one proposed in Figure \ref{fig:kfpc}.

Aligned with the error definition discussed in the previous sections, we introduce the projection and measurement error nodes, \mbox{$\epsilon_y \triangleq \Sigma_z^{-1} (y_{t+1} - C\mu_{t+1})$} and \mbox{$\epsilon_x \triangleq \Sigma_{x_{t+1}}^{-1} (\mu_{t+1} - A\mu_t)$}, respectively. 
Proceeding by computing the gradient of $\tilde{\mathcal{F}}$ with respect to the variational posterior parameters $\mu_{t+1}$, we apply the derivative rules in \eqref{eq:30} and \eqref{eq:32} obtaining:

\begin{equation}
\begin{aligned}
\frac{\partial \tilde{\mathcal{F}}}{\partial \mu_{t+1}} &= -\frac{1}{2} \frac{\partial}{\partial \mu_{t+1}}
 \left[ {\left( (y_{t+1} - C\mu_{t+1})^T \Sigma_{z}^{-1} (y_{t+1} - C\mu_{t+1}) \right)} + { \left( (\mu_{t+1} - A\mu_{t})^T \Sigma_{x_{t+1}}^{-1} (\mu_{t+1} - A\mu_{t}) \right)} \right] \\
&= C^T \Sigma_{z}^{-1} \left( y_{t+1} - C\mu_{t+1}\right) -\Sigma_{x_{t+1}}^{-1} \left( \mu_{t+1} - A\mu_{t}\right) \\
&= C^T\epsilon_y - \epsilon_x.
\end{aligned}
\label{eq:61}
\end{equation}
This results in a gradient with the same form as the one obtained in \eqref{eq:34}, up to the nonlinearity.
Similarly, the matrices $A$ and $C$ can be learned by treating them as parameters and computing the gradient of $\tilde{\mathcal{F}}$ with respect to them. \color{black}
Starting with the state transition matrix $A$, by applying the rule in \eqref{eq:41}, we can write:
\begin{equation}
\begin{aligned}
\frac{\partial \tilde{\mathcal{F}}}{\partial A} &= -\frac{1}{2} \frac{\partial}{\partial A}
 \left[ {\left( (y_{t+1} - C\mu_{t+1})^T \Sigma_{z}^{-1} (y_{t+1} - C\mu_{t+1}) \right)} + { \left( (\mu_{t+1} - A\mu_{t})^T \Sigma_{x_{t+1}}^{-1} (\mu_{t+1} - A\mu_{t}) \right)} \right] \\
&= \epsilon_x\mu_{t}^T.  \\
 \end{aligned}
 \label{eq:62}
\end{equation}
This results in the same gradient obtained for the PCN parameters $\Theta$ in \eqref{eq:38}, up to the linearity of $\mu$. The same applies to the emission matrix $C$:\color{black}
\begin{equation}
\begin{aligned}
\frac{\partial \tilde{\mathcal{F}}}{\partial C} &= -\frac{1}{2} \frac{\partial}{\partial C}
 \left[ {\left( (y_{t+1} - C\mu_{t+1})^T \Sigma_{z}^{-1} (y_{t+1} - C\mu_{t+1}) \right)} + { \left( (\mu_{t+1} - A\mu_{t})^T \Sigma_{x_{t+1}}^{-1} (\mu_{t+1} - A\mu_{t}) \right)} \right] \\
&= \epsilon_y\mu_{t+1}^T.\\
 \end{aligned}
 \label{eq:63}
\end{equation}

Moreover, it is also possible to refine our uncertainty about the system state by optimizing $\tilde{\mathcal{F}}$ with respect to the state covariance matrix, similarly to \eqref{eq:42}. Importantly, the considerations raised in the PCN section regarding covariance updates remain relevant here and represent an active area of research in the PC community. 
\color{black}

This variational treatment of FK under FE optimization thus allows us to reinterpret the problem as a particular case of PC, enhanced with recurrent connections. \color{black} Importantly, unlike KF, this approach results in an approximation of the exact solution. However, it emphasizes how PC with recurrent connections can deal with time series data.  Additionally, KF and its intricate linear algebra equations are unlikely to be implemented by the brain. This perspective instead suggests that it could be approximated by a biologically plausible algorithm, potentially bridging the brain's mechanisms of perception with Bayesian filtering and inference \cite{millidge2024predictive}. \color{black}

\section*{Computational Example}
This section presents a comparative analysis of ANNs trained with PC and BP on standard deep learning tasks common in the PC literature \cite{pinchetti2024benchmarking}, specifically data classification and compression. Experiments focus on standard benchmarking datasets for PC networks \cite{pinchetti2024benchmarking}, such as MNIST, FashionMNIST, and CIFAR10, employing multilayer perceptrons (MLPs) and convolutional neural networks (CNNs) to evaluate performance across deep learning architectures. All models were trained using an NVIDIA RTX A6000 GPU, with model implementation and training performed in Python using the PyTorch framework. For reproducibility of the experiments, the complete implementation setup, including additional details regarding the network architectures and other experiments such as approximating BP gradients using PC, is available on our GitHub repository:
\href{https://github.com/cogsys-tudelft/PredictiveCodingTutorial}{https://github.com/cogsys-tudelft/PredictiveCodingTutorial}.

In data compression tasks, the key difference between the two optimization algorithms lies in their architectural structure. BP autoencoders are composed of two separate networks: an encoder and a decoder. The encoder processes the input to produce a lower-dimensional latent representation, which the decoder uses to attempt reconstructing the original input. The reconstruction error is successively computed and backpropagated through both modules, forcing the encoder to learn compact yet informative representations, and the decoder to accurately reconstruct the input. In contrast, PC achieves the same objective using a single network, as illustrated in Figure~\ref{fig:pcnet2}, effectively acting as an autoencoder folded onto itself. During the PC inference phase, the input is clamped to the bottom-layer neurons, and the network’s neural activity is iteratively updated to minimize the FE according to \eqref{eq:34}. Once convergence is reached, the top layer encodes the compressed representation of the input. The PC decoder functionality is then executed by computing the generative forward pass from this latent code, resulting in the reconstructed input. Importantly, during training, the PC generative parameters are also updated after the inference phase, reducing the mismatch between the reconstruction and the original input. At test time, only the inference phase is performed.  In classification tasks, both BP and PC share the same network architecture and size. Here, the input is clamped to the top layer of the PC network, and the model infers the corresponding label. Once a prediction is made, the true label is clamped to the output layer. Thus, in classification settings, both the first and last layers of a PC network are clamped to fixed values. After clamping the target label, PC inference phase proceeds to propagate eventual feedback errors. Upon convergence, the generative parameters are updated to minimize the FE, as in the compression setting.

\begin{table}[h]
    \centering
    \renewcommand{\arraystretch}{1.2}
    \makebox[0pt][c]{%
    \scalebox{0.9}{%
    \begin{tabular}{|c|c|c|c|c|c|c|c|}
        \hline
        Task & Dataset & Model & Test Metric & Epoch Time (s) & Gen. Params & Var. Params & FLOPs \\
        \hline
        \multirow{6}{*}{\rotatebox{90}{\textbf{Compression}}}
            & \multirow{2}{*}{MNIST} 
                & BP & $(9.39 \pm 0.06) \cdot 10^{-3}$ & $7.01 \pm 0.10$ & 652k & -- & 4M \\
            &   & PC & $(1.10 \pm 0.15) \cdot 10^{-2}$ & $17.34 \pm 0.24$ & 326k & 448 & 138M \\
            \cline{2-8}
            & \multirow{2}{*}{F-MNIST} 
                & BP & $(1.34 \pm 0.14) \cdot 10^{-2}$ & $7.00 \pm 0.19$ & 652k & -- & 4M \\
            &   & PC & $(1.34 \pm 0.33) \cdot 10^{-2}$ & $17.56 \pm 0.57$ & 326k & 448 & 138M \\
            \cline{2-8}
            & \multirow{2}{*}{CIFAR10} 
                & BP & $(1.03 \pm 0.34) \cdot 10^{-2}$ & $8.09 \pm 0.13$ & 2,173k & -- & 11M \\
            &   & PC & $(1.29 \pm 0.61) \cdot 10^{-2}$ & $39.47 \pm 0.55$ & 1,088k & 12k & 226M \\
        \hline
        \multirow{6}{*}{\rotatebox{90}{\textbf{Classification}}} 
            & \multirow{2}{*}{MNIST} 
                & BP & $98.31\% \pm 0.07\,\text{p.p.}$ & $11.64 \pm 0.75$ & 135k & -- & 0.74M \\
            &   & PC & $98.12\% \pm 0.05\,\text{p.p.}$ & $17.40 \pm 1.20$ & 135k & 384 & 5.2M \\
            \cline{2-8}
            & \multirow{2}{*}{F-MNIST} 
                & BP & $89.46\% \pm 0.15\,\text{p.p.}$ & $10.94 \pm 0.22$ & 135k & -- & 0.74M \\
            &   & PC & $88.82\% \pm 0.08\,\text{p.p.}$ & $16.61 \pm 0.27$ & 135k & 384 & 5.2M \\
            \cline{2-8}
            & \multirow{2}{*}{CIFAR10} 
                & BP & $77.56\% \pm 0.45\,\text{p.p.}$ & $11.95 \pm 0.09$ & 4,762k & -- & 48.7M \\
            &   & PC & $77.89\% \pm 0.14\,\text{p.p.}$ & $28.61 \pm 0.06$ & 4,762k & 43k & 313M \\
        \hline
    \end{tabular}%
    }%
    }
    \caption{Comparison of BP and PC across compression and classification tasks.}
    \label{tab:combined_performance}
\end{table}

Table \ref{tab:combined_performance} provides a comparative summary of the performance and resource utilization of both PC and BP models, averaged across five independent runs for each of the three datasets. For MNIST and FashionMNIST, in both classification and compression tasks, an MLP network is used. For CIFAR10, a convolutional architecture is employed. While performance in the classification task is evaluated using accuracy, the metric of interest for the compression task is the MSE loss.The FLOPs estimate refers to the total amount of floating  point operations required by each algorithm to complete a full parameters update. For BP, this includes the forward pass, loss computation, error backpropagation, and parameters update. For PC, it accounts also for the inference phase. The latter was performed over 35 steps for the compression task and over 10 steps in the classification one. FLOPs estimates were obtained using the PyTorch Profiler.
Table \ref{tab:combined_performance} highlights how, in the compression task, PC requires only half the generative parameters compared to BP, while maintaining comparable performance and relying only on local and bio-plausible learning rules. However, PC consistently requires greater computational resources due to its inner optimization process, leading to more FLOPs and longer training times compared to BP. The classification section in Table \ref{tab:combined_performance} shows how PC needs exactly the same number of generative parameters as BP, while having also a small percentage of additional variational parameters. Furthermore, even if in classification settings the inference phase usually requires fewer steps, it still introduces a non-negligible overhead in terms of computation. While PC is competitive with BP for small networks and datasets, it tends to underperform relative to BP when applied to larger datasets and more complex architectures.
 This limitation is discussed in detail in \cite{pinchetti2024benchmarking}, where the authors present a comprehensive comparison of state-of-the-art PC algorithms across various deep learning architectures and datasets, benchmarked against their BP counterparts. In particular, for very deep networks and more complex datasets, BP consistently outperforms PC. These findings highlight the scalability limitations currently affecting the the latter. Although the framework offers appealing features from both computational and biological perspectives, improving its efficiency and performance on large-scale tasks remains an open and actively researched challenge.

\color{black}

\section*{Acknowledgments}
The authors would like to thank Martin Lefebvre for constructive feedback on the manuscript. This publication is part of the project SynergAI with file number NGF.1607.22.010 of the AiNed Fellowship research programme, which is financed by the Dutch Research Council (NWO) and the Microelectronics Department of TU Delft.

\section*{Authors}
\textbf{\textit{Davide Casnici}} (d.casnici@tudelft.nl) received his B.Sc.~degree in Information Engineering from the University of Modena and Reggio Emilia, Italy, in 2021. He successively obtained his M.Sc.~degree in Artificial Intelligence from the University of Lugano, Switzerland, in 2023. Since 2023, he has been a Ph.D.~student at Delft University of Technology, the Netherlands. His research interests include neuromorphic computing, bio-inspired AI, machine learning, statistics, and information theory.
\newline
\newline
\textbf{\textit{Charlotte Frenkel}} (c.frenkel@tudelft.nl)  received her Ph.D. from Université catholique de Louvain in 2020 and was a post-doctoral researcher at the Institute of Neuroinformatics, UZH, and ETH Zürich, Switzerland. She is now an Assistant Professor at Delft University of Technology, The Netherlands. Her research aims at bridging the bottom-up (bio-inspired) and top-down (engineering-driven) design approaches toward neuromorphic intelligence, with a focus on digital neuromorphic processor design, embedded machine learning, and brain-inspired on-device learning. She co-leads the NeuroBench initiative and the Edge AI Foundation working group on neuromorphic engineering.
\newline
\newline
\textbf{\textit{Justin Dauwels}} (j.h.g.dauwels@tudelft.nl) obtained his PhD degree in electrical engineering at the Swiss Polytechnical Institute of Technology (ETH) in Zurich in December 2005. Moreover, he was a postdoctoral fellow at the RIKEN Brain Science Institute (2006-2007) and a research scientist at the Massachusetts Institute of Technology (2008-2010). He is an Associate Professor in AI at TU Delft, co-director of the Safety and Security Institute, and scientific lead of the Model-Driven Decisions Lab, collaborating with the Netherlands police. His research focuses on machine learning, generative AI, and their applications to autonomous systems, as well as the analysis of human behavior and physiology. His academic lab has fostered four startups spanning industries such as health tech and autonomous vehicles.

\bibliographystyle{unsrt}
\bibliography{main}

\begin{thebibliography}{10}

\bibitem{elias1955predictiveI}
P.~Elias.
\newblock Predictive coding--i.
\newblock {\em IRE Transactions on Information Theory}, 1(1):16--24, 1955.

\bibitem{rao1999predictive}
R.~PN. Rao and D.~H. Ballard.
\newblock Predictive coding in the visual cortex: a functional interpretation of some extra-classical receptive-field effects.
\newblock {\em Nature neuroscience}, 2(1):79--87, 1999.

\bibitem{friston2005theory}
K.~Friston.
\newblock A theory of cortical responses.
\newblock {\em Philosophical transactions of the Royal Society B: Biological sciences}, 360(1456):815--836, 2005.

\bibitem{friston2009predictive}
K.~Friston and S.~Kiebel.
\newblock Predictive coding under the free-energy principle.
\newblock {\em Philosophical transactions of the Royal Society B: Biological sciences}, 364(1521):1211--1221, 2009.

\bibitem{whittington2017approximation}
J.~CR. Whittington and R.~Bogacz.
\newblock An approximation of the error backpropagation algorithm in a predictive coding network with local hebbian synaptic plasticity.
\newblock {\em Neural computation}, 29(5):1229--1262, 2017.

\bibitem{millidge2022predictive}
B.~Millidge, A.~Tschantz, and C.~L. Buckley.
\newblock Predictive coding approximates backprop along arbitrary computation graphs.
\newblock {\em Neural Computation}, 34(6):1329--1368, 2022.

\bibitem{song2020can}
Y.~Song, T.~Lukasiewicz, Z.~Xu, and R.~Bogacz.
\newblock Can the brain do backpropagation?---exact implementation of backpropagation in predictive coding networks.
\newblock {\em Advances in neural information processing systems}, 33:22566--22579, 2020.

\bibitem{millidge2021neural}
B.~Millidge, A.~Tschantz, Anil Seth, and C.~Buckley.
\newblock Neural kalman filtering.
\newblock {\em arXiv preprint arXiv:2102.10021}, 2021.

\bibitem{millidge2024predictive}
B.~Millidge, M.~Tang, M.~Osanlouy, N.~S. Harper, and R.~Bogacz.
\newblock Predictive coding networks for temporal prediction.
\newblock {\em PLOS Computational Biology}, 20(4):e1011183, 2024.

\bibitem{marsh2013introduction}
C.~Marsh.
\newblock Introduction to continuous entropy.
\newblock {\em Department of Computer Science, Princeton University}, 1034, 2013.

\bibitem{zahid2023predictive}
U.~Zahid, Q.~Guo, and Z.~Fountas.
\newblock Predictive coding as a neuromorphic alternative to backpropagation: A critical evaluation.
\newblock {\em Neural Computation}, 35(12):1881--1909, 2023.

\bibitem{pinchetti2024benchmarking}
Luca Pinchetti, Chang Qi, Oleh Lokshyn, Gaspard Olivers, Cornelius Emde, Mufeng Tang, Amine M'Charrak, Simon Frieder, Bayar Menzat, Rafal Bogacz, et~al.
\newblock Benchmarking predictive coding networks--made simple.
\newblock {\em arXiv preprint arXiv:2407.01163}, 2024.

\end{thebibliography}

\newpage

\end{document}